\documentstyle[twoside,palatino,fullname,psfig]{article}
\begin{document}

\newtheorem{arule}{Rule}
%==============================
% Commands for formatting stuff
%==============================

% for doing lambda notation with variable x
\newcommand{\ld}[2]{$\lambda$#1$\hspace{.08em}\cdot$\hspace{.12em}#2}
\newcommand{\x}[1]{\ld{X}{#1}}
\newcommand{\xy}[1]{$\lambda$X$\hspace{.08em}\cdot$\hspace{.08em}{\ld{Y}{#1}}}

% for putting `model stuff' in \tt font
\newcommand{\te}{\sl}
\newcommand{\s}[1]{{\sl #1\/}}

%-------------------------------------
% commands for formatting plan schemas
%-------------------------------------
% for formatting plan schemas
\newlength{\planlen}
\addtolength{\planlen}{\textwidth}
\addtolength{\planlen}{-9.5em}
\newcommand{\pplan}[3]
{
   \te
   \begin{tabular}{|ll|}
     \hline
     \hspace{6.6em} & \hspace{\planlen} \vspace{-.7em}
     #3
     \vspace{-.9em} \\ & \\
     \hline
   \end{tabular} 
   \vspace{.2em}   
   \caption{#2}
   \label{plan:#1}
}

\newcommand{\planhead}[1]{\\ {\rm Header:} & #1}
\newcommand{\planwhe}[1]{\\ {\rm Where:} & #1}
\newcommand{\plancon}[1]{\\ {\rm Constraint:} & #1}
\newcommand{\plandec}[1]{\\ {\rm Decomposition:} & #1}
\newcommand{\planeff}[1]{\\ {\rm Effect:} & #1}
\newcommand{\plancont}{\hspace{1.6em}}

%---------------------
% for formatting rules (used in section 5)
%---------------------

\newlength{\ruleant}
\newlength{\infermain}
\newlength{\rulemain}
\newcounter{rulectr}

\newcommand{\crule}[2]{
  \setlength{\ruleant}{\textwidth}
  \addtolength{\ruleant}{-3em}
  {\samepage\sl
  \begin{arule} \mbox{} \\
      #1 $\Longleftarrow$ \\
      \hspace*{\parindent} \parbox[t]{\ruleant}{#2}
  \end{arule}}
}

%--------------------------------------------------------------
% for formatting beliefs and goals that are adopted or inferred
%--------------------------------------------------------------

% an `environment' for beliefs, and goals that are adopted and
%  surface actions that are generated
\newcommand{\infers}[1]
{{\par
 \te
 \vspace{\baselineskip}
 \setlength{\infermain}{\textwidth}
 \addtolength{\infermain}{-1.6pc}
 \hfill\parbox{\infermain}{#1}
 \\
}}

% a command for putting a number on the right of a belief or goal that
% is adopted or inferred.
\setcounter{rulectr}{10}
\newcommand{\infer}[2]{
  \refstepcounter{rulectr}
  \label{#2}
  {\te #1} \hfill (\therulectr) \\
}

%-------------------------
% for formatting dialogues
%-------------------------
\newcommand{\dialogue}[1]
{
   \begin{tabbing}
   \hspace*{1.5em} \= (5.55) \= {\bf B:} \= $^9$ \= \kill
   #1
   \end{tabbing}\vspace{-1em}
}

\newcommand{\dialine}[4]
{
  \>
  #1 \>
  {\bf #2:} \>
  $^#3$ \>
  \parbox[t]{30em}{#4 \vspace*{.5em} } \\
}

\title{Collaborating on Referring Expressions\thanks{To appear in {\em Computational Linguistics}, Volume 21-3, 1995}}
\author{
\begin{tabular}{cc}
Peter A. Heeman                & Graeme Hirst  \\
Department of Computer Science & Department of Computer Science \\
University of Rochester\thanks{This research was began at the Department of Computer Science,
University of Toronto, in the first author's MSc thesis under the
supervision of the second author.} & University of Toronto \\
Rochester, New York            & Toronto, Canada \\
14627			       & M5S 1A4 \\
heeman@cs.rochester.edu        & gh@cs.toronto.edu
\end{tabular}
}
\date{}

\maketitle
\begin{center}
Technical Report 435 \\
Department of Computer Science \\
University of Rochester \vspace{1em} \\
Revised April 1995
\end{center}
\vspace{2em}
\begin{abstract}
\noindent 
This paper presents a computational model of
how conversational participants collaborate in order to make a
referring action successful.  The model is based on the view of
language as goal-directed behavior.  We propose that the content of
a referring expression can be accounted for by the planning
paradigm.  Not only does this approach allow the processes of
building referring expressions and identifying their referents to be
captured by plan construction and plan inference, it also allows us
to account for how participants clarify a referring expression by
using meta-actions that reason about and manipulate the plan
derivation that corresponds to the referring expression.  To account
for how clarification goals arise and how inferred clarification plans
affect the agent, we propose that the agents are in a certain state
of mind, and that this state includes an intention to achieve the goal of
referring and a plan that the agents are currently considering.  It
is this mental state that sanctions the adoption of goals and the
acceptance of inferred plans, and so acts as a link between
understanding and generation.
\end{abstract}
\cleardoublepage

%=====================
\section{Introduction}
%=====================

People are goal oriented and can plan courses of actions to
achieve their goals.  But sometimes they might lack the knowledge
needed to formulate a plan of action, or some of the actions that they
plan might depend on coordinating their activity with other agents.
How do they cope?  One way is to work together, or {\it collaborate\/}, in
formulating a plan of action with other people who are involved in the
actions or who know the relevant information.

Even in the apparently simple linguistic task of referring, in an
utterance, to some object or idea can involve exactly this kind of
activity: a collaboration between the speaker and the hearer.  The
speaker has the goal of the hearer identifying the object that the
speaker has in mind.  The speaker attempts to achieve this goal by
constructing a description of the object that she thinks will enable
the hearer to identify it.  But since the speaker and the hearer will
inevitably have different beliefs about the world, the hearer might
not be able to identify the object.  Often, when the hearer cannot do
so, the speaker and hearer collaborate in making a new referring
expression that accomplishes the goal.

This paper presents a computational model of how a conversational
participant collaborates in making a referring action successful.  We
use as our basis the model proposed by Clark and Wilkes-Gibbs
\shortcite {clark-wilkesgibbs:86}, which gives a
descriptive account of the conversational moves that participants
make when collaborating upon a referring expression.  We cast their
work into a model based on the planning paradigm.

We propose that referring expressions can be represented by plan
derivations, and that plan construction and plan inference can be used
to generate and understand them.  Not only does this approach allow
the processes of {\it building\/} referring expressions and
{\it identifying\/} their referents to be captured in the planning
paradigm, it also allows us to use the planning paradigm to account
for how participants {\it clarify\/} a referring expression.  In this
case, we use meta-actions that encode how a plan derivation
corresponding to a referring expression can be reasoned about and
manipulated.

To complete the picture, we also need to account for the fact that the
conversants are {\it collaborating\/}.  We propose that the agents are in
a mental state that includes not only an intention to achieve
the goal of the collaborative activity but also a plan that the
participants are currently considering.  In the
case of referring, this will be the plan derivation that corresponds to
the referring expression.  This plan is in the common ground of the
participants, and we propose rules that are sanctioned by the mental
state both for {\it accepting\/} plans that clarify the current plan, and
for {\it adopting\/} goals to do likewise.  The acceptance of a
clarification results in the current plan
being updated.  So, it is these rules that specify how plan inference
and plan construction affect and are affected by the mental state of
the agent.  Thus, the mental state, together with the rules, provides the
link between these two processes.  An important consequence of our
proposal is that the current plan need not allow the successful
achievement of the goal.  Likewise, the clarifications that agents propose
need not result in a successful plan in order for them to be accepted.

As can be seen, our approach consists of two tiers.  The first tier is
the planning component, which accounts for how utterances are both
understood and generated.  Using the planning paradigm has several
advantages: it allows both tasks to be captured in a single paradigm
that is used for modeling general intelligent behavior; it allows more
of the content of an utterance to be accounted for by a uniform
process; and only a single knowledge source for referring expressions
is needed instead of having this knowledge embedded in special
algorithms for each task.  The second tier accounts for the
collaborative behavior of the agents: how they adopt goals and
coordinate their activity.  It provides the link between the mental
state of the agent and the planning processes.

In accounting for how agents collaborate in making a
referring action, our work aims to make the following contributions
to the field.  First, although much work has been done on how agents
request clarifications, or respond to such requests, little attention
has been paid to the collaborative aspects of clarification discourse.
Our work attempts a plan-based formalization
of what linguistic collaboration is, both in terms of the goals and
intentions that underlie it and the surface speech acts that result
from it.  Second, we address the act of referring and show how it can
be better accounted for by the planning paradigm.  Third, previous
plan-based linguistic research has concentrated on either construction
or understanding of utterances, but not both.  By doing both, we will
give our work generality in the direction of a complete model of the
collaborative process.  Finally, by using Clark and Wilkes-Gibbs's
model as a basis for our work, we aim not only to add support to their
model, but gain a much richer understanding of the subject.

In order to address the problem that we have set out, we have limited
the scope of our work.  First, we look at referring expressions in
isolation, rather than as part of a larger speech act.  Second, we
assume that agents have mutual knowledge of the mechanisms of
referring expressions and collaboration.  Third, we deal with objects
that both the speaker and hearer know of, though they might have
different beliefs about what propositions hold for these objects.
Fourth, as the input and the output to our system, we use
representations of surface speech actions, not natural language
strings.  Finally, although belief revision is an important part of
how agents collaborate, we do not explicitly address this.

%=============================================
\section{Referring as a Collaborative Process}
%=============================================

Clark and Wilkes-Gibbs \shortcite{clark-wilkesgibbs:86} investigated
how participants in a conversation collaborate in making a referring
action successful.  They conducted experiments in which participants
had to refer to objects---tangram patterns---that are difficult to
describe.  They found that typically the participant trying to
describe a tangram pattern would present an initial referring
expression.  The other participant would then pass judgment on it,
either {\it accepting\/} it, {\it rejecting\/} it, or {\it postponing\/} his
decision.  If it was rejected or the decision postponed, then one
participant or the other would {\it refashion\/} the referring expression.
This would take the form of either repairing the expression by
correcting speech errors, {\it expanding\/} it by adding further
qualifications, or {\it replacing\/} the original expression with a new
expression. The referring expression that results from this is then
judged, and the process continues until the referring expression is
acceptable enough to the participants for current purposes.  This
final expression is contributed to the participants' common ground.

Below are two excerpts from Clark and Wilkes-Gibbs's experiments that
illustrate the acceptance process.
\dialogue{
  \dialine{(2.1)}{A}{1}{Um, third one is the guy reading with, holding his
	book to the left.}
  \dialine{}{B}{2}{Okay, kind of standing up?}
  \dialine{}{A}{3}{Yeah.}
  \dialine{}{B}{4}{Okay.}
}
In this dialogue, person A makes an initial presentation in line~1.
Person B postpones his decision in line~2 by voicing a {\it tentative
``okay''\/}, and then proceeds to refashion the referring expression,
the result being ``the guy reading, holding his book to the left,
kind of standing up.''  A accepts the new expression in line~3, and B
signals his acceptance in line~4.
\dialogue{
  \dialine{(2.2)}{A}{1}{Okay, and the next one is the person that
	looks like they're carrying something and it's sticking out
	to the left.  It looks like a hat that's upside down.}
  \dialine{}{B}{2}{The guy that's pointing to the left again?}
  \dialine{}{A}{3}{Yeah, pointing to the left, that's it! (laughs)}
  \dialine{}{B}{4}{Okay.}
}
In the second dialogue, B implicitly rejects A's initial
presentation by replacing it with a new referring expression in
line~2, ``the guy that's pointing to the left again.''
A then accepts the refashioned referring expression in line~3.

Below, we give an algorithmic
interpretation of Clark and Wilkes-Gibbs's collaborative model, where
{\bf present}, {\bf judge}, and {\bf refashion} are the conversational moves
that the participants make, and {\it ref\/}, {\it re\/}, and {\it judgment\/} are
variables that represent the referent, the current referring
expression, and its judgment, respectively.  (Since the conversational
moves update the referring expression and its judgment, they are
presented as functions.)
\infers{{\it re\/} $=$ {\bf present(}{\it ref\/}{\bf )} \\
{\it judgment\/} $=$ {\bf judge(}{\it ref,re\/}{\bf )} \\
{\sl while (}{\it judgment\/} $\not = $ {\bf accept}{\sl )} \\
\hspace*{1em} {\it re\/} $=$ {\bf refashion(}{\it ref,re\/}{\bf )} \\
\hspace*{1em} {\it judgment\/} $=$ {\bf judge(}{\it ref,re\/}{\bf )} \\
{\sl end-while} \\}
The algorithm illustrates how the collaborative activity progresses by
the participants judging and refashioning the previously proposed
referring expression.\footnote{For simplicity, we have not shown the
change in speakers between refashionings and judgments.} In fact, we
can see that the {\it state\/} of the process is characterized by the
current referring expression, {\it re\/}, and the judgment of it, {\it
judgment\/}, and that this state must be part of the common ground of
the participants.  The algorithm also illustrates how the model of
Clark and Wilkes-Gibbs minimizes the distinction between the roles of
the person who initiated the referring expression and the person who
is trying to identify it.  Both have the same moves available to them,
for either can judge the description and either can refashion it.
Neither is controlling the dialogue, they are simply collaborating.

In later work, Clark and Schaefer \shortcite {clark-schaefer:89}
propose that ``each part of the acceptance phase is itself a
contribution'' (p.~269), and the acceptance of these contributions
depends on whether the hearer ``believes he is understanding well
enough for current purposes'' (p.~267).  Although Clark and Schaefer
use the term {\it contribution\/} with respect to the discourse, rather
than the collaborative effort of referring, their proposal is still
relevant here: judgments and refashionings are contributions to the
collaborative effort and are subjected to an acceptance process, with
the result being that once they are accepted, the state of the
collaborative activity is updated.  So, what constitutes grounds for
accepting a judgment or clarification?  In order to be consistent with
Clark and Wilkes-Gibbs' model, we can see that if one agent finds the
current referring expression problematic, the other must accept that
judgment.  Likewise, if one agent proposes a referring expression,
through a refashioning, the other must accept the refashioning.

%==============================
\section{Referring Expressions}
%==============================

%----------------------------------
\subsection{Planning and Referring}
%----------------------------------

By viewing language as action, the planning paradigm can be applied to
natural language processing.  The actions in this case are {\it speech
acts\/} \cite {austin:62,searle:69}, and include such things as
promising, informing, and requesting.  Cohen and Perrault \shortcite
{cohen-perrault:79} developed a system that uses plan construction to
map an agent's goals to speech acts, and Allen and Perrault
\shortcite{allen-perrault:80} use plan inference to understand an
agent's plan from its speech acts.  By viewing it as action
\cite{searle:69}, referring can be incorporated into a planning model.
Cohen's model \shortcite {cohen:81} planned requests that the hearer
identify a referent, whereas Appelt \shortcite {appelt:85j} planned
{\it concept activations\/}, a generalization of referring actions.

Although acts of reference have been incorporated into plan-based
models, determining the content of referring expressions hasn't been.
For instance, in Appelt's model, concept activations can be achieved
by the action \s{describe}, which is a primitive, not further
decomposed.  Rather, this action has an associated procedure that
determines a description that satisfies the preconditions of
\s{describe}.  Such special procedures have been the mainstay for
accounting for the content of referring expressions, both in
constructing and in understanding them, as exemplified by Dale
\shortcite{dale:89}, who chose descriptors on the basis of their
discriminatory power, Ehud Reiter \shortcite{e.reiter:90}, who focused
on avoiding misleading conversational implicatures when generating
descriptions, and Mellish \shortcite{mellish:85:book}, who used a
constraint satisfaction algorithm to identify referents.

Our work follows the plan-based approach to language generation and
understanding.  We extend the earlier approaches of Cohen and
Appelt by accounting for the content of the description at the
planning level.  This is done by having surface speech actions for
each component of a description, plus a surface speech action that
expresses a speaker's intention to refer.  A referring action is
composed of these primitive actions, and the speaker utters
them in her attempt to refer to an object.

These speech actions are the building blocks that referring
expressions are made from.  Acting as the mortar are intermediate
actions, which have constraints that the plan construction and plan
inference processes can reason about.  These constraints encode the
knowledge of how a description can allow a hearer to identify an
object.  First, the constraints express the conditions under which an
attribute can be used to refer to an object; for instance, that it be
mutually believed that the object has a certain property
\cite{clark-marshall:81,perrault-cohen:81,nadathur-joshi:83}.  Second, 
the constraints keep track of which objects could be believed to be
the referent of the referring expression.  Third, the constraints
ensure that a sufficient number of surface speech actions are added so
that the set of candidates associated with the entire referring
expression consists of only a single object, the referent.  These
constraints enable the speaker to construct a referring
expression that she believes will allow the hearer to identify
the referent.  As for the hearer, the explicit encoding of the adequacy
of referring expressions allows referent identification to fall out of
the plan inference process.

Our approach to treating referring as a plan in which surface
speech actions correspond to the components of the description
allows us to capture how participants collaborate in building a
referring expression.  Plan repair techniques can be used to refashion
an expression if it is not adequate, and clarifications can refer to the
part of the plan derivation that is in question or is being repaired.
Thus we can model a collaborative dialogue in terms of the changes
that are being made to the plan derivation.

The referring expression plans that we propose are not simply data
structures, but are mental objects that agents have beliefs about
\cite {pollack:90}.  The plan derivation expresses beliefs of the
speaker: how actions contribute to the achievement of the goal, and
what constraints hold that will allow successful
identification.\footnote {Since we assume that the agents have mutual
knowledge of the action schemas and that agents can execute surface
speech actions, we do not consider beliefs about generation or about
the executability of primitive actions.} So plan construction reasons
about the beliefs of the agent in constructing a referring plan;
likewise, plan inference, after hypothesizing a plan that is
consistent with the observed actions, reasons about the other
participant's (believed) beliefs in satisfying the constraints of the
plan.  If the hearer is able to satisfy the constraints, then he will
have understood the plan and be able to identify the referent, since a
term corresponding to it would have been instantiated in the inferred
plan.  Otherwise, there will be an action that includes a
constraint that is unsatisfiable, and the hearer construes the action as
being in error.  (We do not reason about how the error affects the
satisfiability of the goal of the plan nor use the error to revise the
beliefs of the hearer.)

%===================================
\subsection{Vocabulary and Notation}
%===================================

Before we present the action schemas for referring expressions, we
need to introduce the notation that we use.  Our terminology for
planning follows the general literature.\footnote {See the
introductory chapter of Allen, Hendler, and Tate \shortcite
{readings-in-planning:90} for an overview of planning.} We use the
terms {\it action schema\/}, {\it plan derivation\/}, {\it plan
construction\/}, and {\it plan inference\/}.  An action schema consists
of a {\it header\/}, {\it constraints\/}, a {\it decomposition\/}, and an
{\it effect\/}; and it encodes the constraints under which an effect can
be achieved by performing the steps in the decomposition.  A plan
derivation is an instance of an action that has been recursively
expanded into primitive actions---its {\it yield\/}.  Each component in
the plan---the action headers, constraints, steps, and effects---are
referred to as {\it nodes\/} of the plan, and are given names so as to
distinguish two nodes that have the same content.  Finally, plan
construction is the process of finding a plan derivation whose yield
will achieve a given effect, and plan inference is the process of
finding a plan derivation whose yield is a set of observed primitive
actions.

The action schemas make use of a number of predicates and these are
defined in Table~\ref {table:t3_table}.  We adopt the Prolog
convention that variables begin with an upper-case letter, and all
predicates and constants begin with a lower-case letter.  Two
constants that need to be mentioned are \s{system} and \s{user}.  The
first denotes the agent that we are modeling, and the latter, her
conversational partner.  Since the action schemas are used for both
constructing the plans of the \s{system}, and inferring the plans of
the \s{user}, it is sometimes necessary to refer to the speaker or
hearer in a general way.  For this we use the propositions \s{speaker(Speaker)} and \s{hearer(Hearer)}.  These instantiate the
variables \s{Speaker} and \s{Hearer} to \s{system} or \s{user};
which is which depends on whether the rule is being used for plan
construction or plan inference.  These propositions are included as
constraints in the action schemas as needed.
\begin{table}

{\noindent\bf Belief}
\begin{description}
\item[\s{bel(Agt,Prop)}:]
\s{Agt} believes that \s{Prop} is true.

\item[\s{bmb(Agt1,Agt2,Prop)}:]
\s{Agt1} believes that it is mutually believed between
himself and \s{Agt2} that \s{Prop} is true.

\item[\s{knowref(Agt1,Agt2,Ent,Obj)}:]
\s{Agt1} knows the referent that \s{Agt2} associates
with the discourse entity \s{Ent} \cite{webber:83}, which \s{Agt1}
believes to be \s{Obj}.  (Proving this proposition with \s{Ent}
unbound will cause a unique identifier to be created for \s{Ent}.)
\end{description}

{\noindent\bf Goals and Plans}
\begin{description}
\item[\s{goal(Agt,Goal)}:]
\s{Agt} has the goal \s{Goal}.  Agents act to make their
goals true.

\item[\s{plan(Agt,Plan,Goal)}:]
%% Graeme: added `the plan derivation'
\s{Agt} has the goal of \s{Goal} and has adopted the plan
derivation \s{Plan} as a means to achieve it.  The agent believes
that each action of \s{Plan} contributes to the goal, but not
necessarily that all of the constraints hold; in order words, the plan
must be coherent but not necessarily valid \cite[p.~94]{pollack:90}.

\item[\s{content(Plan,Node,Content)}:]
The node named by \s{Node} in \s{Plan} has content
\s{Content}.

\item[\s{yield(Plan,Node,Actions)}:]
The subplan rooted at \s{Node} in \s{Plan} has a yield of the
primitive actions \s{Actions}.

\item[\s{achieve(Plan,Goal)}:]
Executing \s{Plan} will cause \s{Goal} to be true.

\item[\s{error(Plan,Node)}:]
\s{Plan} has an error at the action labeled \s{Node}.
Errors are attributed to the action that contains the failed
constraint.  This predicate is used to encode an agent's belief about
an invalidity in a plan.
\end{description}

{\noindent\bf Plan Repair}
\begin{description}
\item[\s{substitute(Plan,Node,NewAction,NewPlan)}:]
Undo all variable bindings in \s{Plan} (except those in primitive
actions, and then substitute the action header \s{NewAction} into \s{Plan} at \s{Node}, resulting in the partial
plan \s{NewPlan}.

\item[\s{replan(Plan,Actions)}:]
Complete the partial plan \s{Plan}.  \s{Actions} are the
primitive actions that are added to the plan.

\item[\s{replace(Plan,NewPlan)}:]
The plan \s{NewPlan} replaces \s{Plan}.
\end{description}

{\noindent\bf Miscellaneous}
\begin{description}
\item[\s{subset(Set,Lambda,Subset)}:]
Compute the subset, \s{Subset}, of \s{Set} that satisfies the lambda
expression \s{Lambda}.

\item[\s{modifier-absolute-pred(Pred)}:] \s{Pred} is a 
predicate that an object can be described in terms of.  Used by the
\s{modifier-absolute} schema given in Figure~\ref{plan:modifier-abs}.

\item[\s{modifier-relative-pred(Pred)}:]  \s{Pred} is a
predicate that describes the relationship between two objects.  Used
by the \s{modifier-relative} schema given in Figure~\ref
{plan:modifier-rel}.

\item[\s{pick-one(Object,Set)}:] Pick one object, \s{Object},
of the members of \s{Set}.

\item[\s{speaker(Agt)}:] \s{Agt} is the current speaker.

\item[\s{hearer(Agt)}:] \s{Agt} is the current hearer.
\end{description}
\hrule
\caption{\label{table:t3_table}Predicates and Actions}

\end{table}

%==========================
\subsection{Action Schemas}
%==========================

This section presents action schemas for referring expressions.  (We
omit discussion of actions that account for superlative adjectives,
such as ``largest'', that describe an object relative to the set of
objects that match the rest of the description.  A full presentation
is given by Heeman~\shortcite{heeman:91:thesis}.)

As we mentioned, the action for referring, called \s{refer}, is
mapped to the surface speech actions through the use of intermediate
actions and plan decomposition.  All of the reasoning is done in the
\s{refer} action and the intermediate actions, so no constraints or
effects are included in the surface speech actions.

We use three surface speech actions.  The first is
\s{s-refer(Entity)}, which is used to express the speaker's intention to
refer.  The second is \s{s-attrib(Entity,Predicate)}, and is used for
describing an object in terms of an attribute; \s{Entity} is the
discourse entity of the object, and \s{Predicate} is a lambda
expression, such as \s{\x{category(X,bird)}}, that encodes the
attribute.  The third is \s
{s-attrib-rel(Entity,OtherEntity,Predicate)}, and is used for
describing an object in terms of some other object.  In this case
\s{Predicate} is a lambda expression of two variables, one
corresponding to \s{Entity}, and the other to \s{OtherEntity}; for
instance, \s{\xy{in(X,Y)}}.

%----------------------------
\subsubsection*{Refer Action}
%----------------------------
\label{sec:refer}

The schema for \s{refer} is shown in Figure~\ref {plan:refer}.
\begin{figure}[b]
\pplan{refer}{\s{refer} schema}{
  \planhead{    refer(Entity,Object) }
  \plancon{     knowref(Speaker,Speaker,Entity,Object)}
  \plandec{     s-refer(Entity) \\
                & describe(Entity,Object) }
  \planeff{     bel(Hearer,goal(Speaker,knowref(Hearer,Speaker,Entity,Object))) }
}
\end{figure}%
The \s{refer} action decomposes into two steps: \s{s-refer}, which
expresses the speaker's intention to refer, and \s{describe}, which
accounts for the content of the referring expression (given next).
The effect of \s{refer} is that the hearer should believe that the
speaker has a goal of the hearer knowing the referent of the referring
expression.  The effect has been formulated in this way because we are
assuming that when a speaker has a communicative goal she plans to
achieve the goal by making the hearer recognize it; the effect will be
achieved by the hearer inferring the speaker's plan, regardless of
whether or not the hearer is able to determine the actual referent.
To simplify our implementation, this is the only effect that is stated
for the action schemas for referring expressions.  It corresponds to
the literal goal that Appelt and Kronfeld
\shortcite{appelt-kronfeld:87} propose (whereas the actual
identification is their condition of satisfaction).

%------------------------------------
\subsubsection*{Intermediate Actions}
%------------------------------------

The \s{describe} action, shown in Figure~\ref{plan:describe}, is used
to construct a description of the object through its decomposition
into \s{headnoun} and \s{modifiers}.
The variable \s{Cand} is the candidate set, the set of
potential referents, associated with the head noun that is chosen, and
it is passed to the \s{modifiers} action so that it can ensure
that the rest of the description rules out all of the alternatives.
\begin{figure}[h]
\pplan{describe}{\s{describe} schema}{
  \planhead{    describe(Entity,Object)}
  \plandec{     headnoun(Entity,Object,Cand) \\ &
		modifiers(Entity,Object,Cand)}
}
\end{figure}

The action \s{headnoun}, shown in Figure~\ref {plan:headnoun},
has a single step, \s{s-attrib}, which is the surface
speech action used to describe an object in terms of some predicate,
which for the \s{headnoun} schema, is restricted to the category of
the object.\footnote{Note that several category predications might be
true of an object, and we do not explore which would be best to use,
but see Edmonds \shortcite{edmonds:94:coling} for how preferences can
be encoded.} The schema also has two constraints.  The first ensures
that the referent is of the chosen category and the second determines
the candidate set, \s{Cand}, associated with the head noun that is
chosen.  The candidate set is computed by finding the subset of the
objects in the world that the speaker believes could be referred to by
the head noun---the objects that the speaker and hearer have an
appropriate mutual belief about.
\begin{figure}[htb]
\pplan{headnoun}{\s{headnoun} schema}{
  \planhead{    headnoun(Entity,Object,Cand) }
  \plancon{	world(World) \\ &
		bmb(Speaker,Hearer,category(Object,Category)) \\ &
	        subset(World,\x{bmb(Speaker,Hearer,category(X,Category))},Cand)}
  \plandec{     s-attrib(Entity,\x{category(X,Category)})}
}
\end{figure}

The \s{modifiers}\label{sec:modifiers} action attempts to ensure that
the referring expression that is being constructed is believed by the
speaker to allow the hearer to uniquely identify the referent.  We
have defined \s{modifiers} as a recursive action, with two
schemas.\footnote{We use specialization axioms
\cite{kautz-allen:86:aaai} to map the \s{modifiers} action to the
two schemas: \s{modifiers-terminate} and \s{modifiers-recurse}.}
The first schema, shown in Figure~\ref {plan:modifiers-ter}, is used
to terminate the recursion, and its constraint specifies that only one
object can be in the candidate set.\footnote {In order to distinguish
this action from the primitive actions, it has a step that is marked
\s{null}.}  The second schema, shown in Figure~\ref
{plan:modifiers-rec}, embodies the recursion.  It uses the
\s{modifier} plan, which adds a component to the description and
updates the candidate set by computing the subset of it that satisfies
the new component.  The \s{modifier} plan thus accounts for individual
components of the description.
\begin{figure}[thb]
\pplan{modifiers-ter}{\s{modifiers} schema for terminating the recursion}{
  \planhead{    modifiers-terminate(Entity,Object,Cand) }
  \plancon{     Cand = [Object]}
  \plandec{     null}
}
\end{figure}
\begin{figure}[thb]
\pplan{modifiers-rec}{\s{modifiers} schema for recursing}{
  \planhead{    modifiers-recurse(Entity,Object,Cand) }
  \plandec{     modifier(Entity,Object,Cand,NewCand) \\ &
                modifiers(Entity,Object,NewCand)}
}
\end{figure}

There are two different action schemas for \s{modifier};
one is for absolute modifiers, such as ``black'' and the other is for
relative modifiers, such as ``larger''.  The former is shown in Figure~\ref
{plan:modifier-abs};
it decomposes into the
surface speech action \s{s-attrib} and has a constraint that
determines the new candidate set, \s{NewCand}, by including only the
objects from the old candidate set, \s{Cand}, for which the predicate
could be believed to be true.  The other schema is shown in
Figure~\ref{plan:modifier-rel} and is used for
describing objects in terms of some other object.
It uses the surface speech action \s{s-attrib-rel}\label{sec:s-attrib-rel} and also includes a
step to refer to the object of comparison.
\begin{figure}[hbt]
\pplan{modifier-abs}{\s{modifier} schema for absolute modifiers}{
  \planhead{    modifier-absolute(Entity,Object,Cand,NewCand) }
  \plancon{     modifier-pred(Pred) \\ &
                bmb(Speaker,Hearer,Pred(X)) \\ &
                subset(Cand,\x{bmb(Speaker,Hearer,Pred(X))},NewCand)}
  \plandec{     s-attrib(Entity,Pred)}
}
\end{figure} \mbox{}
\begin{figure}[hbt]
\pplan{modifier-rel}{\s{modifier} schema for relative modifiers}{
  \planhead{    modifier-relative(Entity,Object,Cand,NewCand) }
  \plancon{     modifier-rel-pred(Pred) \\ &
                bmb(Speaker,Hearer,Pred(Object,OtherObject)) \\ &
                subset(Cand,\x{bmb(Speaker,Hearer,Pred(X)(Other))},NewCand)}
v  \plandec{     s-attrib-rel(Entity,OtherEntity,Pred) \\ & 
                refer(OtherEntity,Other)}
}
\end{figure}

%================================================
\subsection{Plan Construction and Plan Inference}
%================================================

The goals that we are interested in achieving are communicative goals.
Since these goals cannot be directly achieved by a plan of action, the
speaker must instead plan actions that will achieve them indirectly,
for instance by planning an utterance that results in the hearer
recognizing her goal.  So, if the speaker wants to achieve \s{Goal},
she will attempt to construct a plan whose effect is
\s{bel(Hearer,goal(Speaker,Goal))}.

%--------------------------------
\subsubsection*{Plan Construction}
%--------------------------------

Given an effect, the plan constructor finds a plan derivation that has
a minimal number of primitive action, that is valid (with respect to
the planning agent's beliefs), and whose root action achieves the
effect.  The plan constructor uses a best-first search strategy,
expanding the derivation with the fewest number of surface speech
actions. The yield of this plan derivation can then be given as input
to a module that generates the surface form of the
utterance.

%-----------------------------
\subsubsection*{Plan Inference}
%-----------------------------

Following Pollack \shortcite{pollack:90}, our plan inference process
can infer plans in which, in the hearer's view, a constraint does not
hold.  In inferring a plan derivation, we first find the set of plan
derivations that account for the primitive actions that were observed,
without regard to whether the constraints hold.  This is done by using
a chart parser that parses actions rather than words \cite
{sidner:85,vilain:90}.  For referring plans that contain
more than one modifier, there will be multiple derivations
corresponding to the order of the modifiers.  We avoid this ambiguity
by choosing an arbitrary ordering of the modifiers for each such plan.

In the second part of the plan inference process, we evaluate each
derivation by attempting to find an instantiation for the variables
such that all of the constraints hold with respect to the hearer's
beliefs about the speaker's beliefs.  It could however be the case
that there is no instantiation, either because this is not the right
derivation or because the plan is based on beliefs not shared by the
speaker and the hearer.  In the latter case, we need to determine
which action in the plan is to blame, so that this knowledge can be
shared with the other participant.

After each derivation has been evaluated, if there is just one valid
derivation, an instantiated derivation whose constraints all hold,
then the hearer will believe that he has understood.  If there is just
one derivation and it is invalid, the action containing the constraint
that is the source of the invalidity is noted.  (We have not explored
ambiguous situations, those in which more than one valid derivation
remains, or, in the absence of validity, more than one invalid
derivation.)

We now need to address how we evaluate a derivation.  In the case
where the plan is invalid, we need to partially evaluate the plan in
order to determine which action contains a constraint that cannot be
satisfied.  However, any instantiation will lead to some constraint
being found not to hold. Care must therefore be taken in finding the
right instantiation so that blame is attributed to the action at
fault.  So, we evaluate the constraints in order of mention in the
derivation, but postpone any constraints that have multiple solutions
until the end.  We have found that this simple approach can find the
instantiation for valid plans and can find the action that is in error
for the others.

To illustrate this, consider the \s{headnoun} action,
which has the following constraints.
\infers{
speaker(Speaker) \\
hearer(Hearer) \\
world(World) \\
bmb(Speaker,Header,category(Object,Category)) \\
subset(World,\x{bmb(Speaker,Hearer,category(X,Category))},Cand) \\}
During the first step, finding the derivation, all co-referential
variables will be unified.  In particular, the variable \s{Category}
will be instantiated from the co-referential variable in the surface
speech action.  The first three constraints have only a single
solution, so they are instantiated.  The fourth constraint contains
\s{Object}.  If there is exactly one object that the system believes
to be mutually believed to be of \s{Category}, then \s{Object} is
instantiated to it.  If there is none, then this constraint is
unsatisfiable, and so the evaluation of this plan stops with this
action marked as being in error, since no object matches this part of
the description.  If there is more than one, then this constraint is
postponed and the evaluator moves on to the \s{subset} constraint.
This constraint has one uninstantiated variable, \s{Cand}, which has
a unique (non-null) solution, namely the candidate set associated with
the head noun.  So, this constraint is evaluated.

The evaluation then proceeds through the actions in the rest of the
plan.  Assuming that no intervening errors are encountered, the
evaluator will eventually reach the constraint on the terminating
instance of \s{modifiers}, \s{Cand = [Object]}, with \s{Cand}
instantiated to a non-null set.  If \s{Cand} contains more than one
object, then this constraint will fail, pinning the blame on the
terminating instance of \s{modifiers} for there not being enough
descriptors to allow the referent to be identified.  Otherwise, the
terminating constraint will be satisfiable, and so \s{Object} will
be instantiated to the single object in the candidate set.  This will
then allow all of the mutual belief constraints that were postponed to
be evaluated, since they will now have only a
single solution.

%=======================
\section{Clarifications}
%=======================

%-----------------------------------
\subsection{Planning and Clarifying}
%-----------------------------------
\label{text:Planning-and-Clarifying}

Clark and Wilkes-Gibbs \shortcite{clark-wilkesgibbs:86} have presented
a model of how conversational participants collaborate in making a
referring action successful (see section 2 above).  Their model
consists of conversational moves that express a judgment of a
referring expression and conversational moves that refashion an
expression.  However, their model is not computational.  They do not
account for how the judgment is made, how the judgment affects the
refashioning, nor the content of the moves.

Following the work of Litman and Allen \shortcite{litman-allen:87} in
understanding clarification subdialogues, we formalize the
conversational moves of Clark and Wilkes-Gibbs as discourse actions.
These discourse actions are meta-actions that take as a parameter a
referring expression plan.  The constraints and decompositions of the
discourse actions encode the conditions under which they can be
applied, how the referring expression derivations can be refashioned,
and how the speaker's beliefs can be communicated to the hearer.  So,
the conversational moves, or clarifications, can be generated and
understood within the planning paradigm.\footnote{We use the term
{\it clarification\/}, since the conversational moves of judging and
refashioning a referring expression can be viewed as clarifying it.}

%--------------------------------------
\subsubsection*{Surface Speech Actions}
%--------------------------------------

An important part of our model is the surface speech actions.  These
actions serve as the basis for communication between the two agents,
and so they must convey the information that is dictated by Clark and
Wilkes-Gibbs's model.  For the judgment plans, we have the surface
speech actions \s{s-accept}, \s{s-reject}, and \s{s-postpone}
corresponding to the three possibilities in their model.  These take
as a parameter the plan that is being judged, and for \s{s-reject}, also a
subset of the speech actions of the referring expression plan.  The
purpose of this subset is to inform the hearer of the surface speech
actions that the speaker found problematic.  So, if the referring
expression was ``the weird creature'', and the hearer couldn't
identify anything that he thought ``weird'', he might say ``what weird
thing'', thus indicating he had problems with the surface speech
action corresponding to ``weird''.

For the refashioning plans, we propose that there is a single surface
speech action, \s{s-actions}, that is used for both replacing a part
of a plan, and expanding it.  This action takes as a parameter the
plan that is being refashioned, and a set of surface speech actions
that the speaker wants to incorporate into the referring expression
plan.  Since there is only one action, if is in uttered in isolation,
it will be ambiguous between a replacement and an expansion; however,
the speech action resulting from the judgment will provide the proper
context to disambiguate its meaning.  In fact, during linguistic
realization, if the two actions are being uttered by the same person,
they could be combined into a single utterance.  For instance, the
utterance ``no, the red one'' could be interpreted as a \s{s-reject}
of the color that was previously used to describe something and an
\s{s-actions} for the color ``red.''

So, as we can see, the surface speech actions for clarifications
operate on components of the plan that is being built, namely the
surface speech actions of referring expression plans.  This is
consistent with our use of plan derivations to represent utterances.
Although we could have viewed the clarification speech actions as acts
of informing \cite{litman-allen:87}, this
would have shifted the complexity into the parameter of the \s{inform}
and it is unclear whether anything would have been gained.  Instead,
we feel that a parser with a model of the discourse and the context
can determine the surface speech actions.\footnote{See Levelt
\shortcite[Chapter 12]{levelt:89:speaking} for how prosody and clue 
words can be used in determining the type of clarification.}
Additionally, it should be easier for the generator to determine an
appropriate surface form.

%-------------------------------
\subsubsection*{Judgment Plans}
%-------------------------------

The evaluation of the referring expression plan indicates whether the
referring action was successful or not.  If it was successful, then
the referent has been identified, and so a goal to communicate this is
input to the plan constructor.  This goal would be achieved by an
instance of \s{accept-plan}, which decomposes into the
surface speech action \s{s-accept}.

If the evaluation wasn't successful, then the goal of communicating
the error is given to the plan constructor, where the error is simply
represented by the node in the derivation that the evaluation failed
at.  There are two reasons why the evaluation could
have failed, either no objects match, or more than one matches.  In
the first case, the referring expression is overconstrained, and the
evaluation would have failed on an action that decomposes into surface
speech actions. In the second case, the referring expression is
underconstrained, and so the evaluation would have failed on the
constraint that specifies the termination of the addition of
modifiers.  In our formalization of the conversational moves, we have
equated the first case to \s{reject-plan} and the second case to
\s{postpone-plan}, and their constraints test for the abovementioned
conditions. The actions \s{reject-plan} and \s{postpone-plan} decompose into the surface speech actions \s{s-reject} and \s{s-postpone}, respectively.

By observing the surface speech action corresponding to the judgment,
the hearer, using plan inference, should be able to derive the
speaker's judgment plan. If the judgment was \s{reject-plan} or \s{postpone-plan}, then the evaluation of the judgment plan should enable
the hearer to determine the action in the referring plan that the
speaker found problematic due to the constraints specified
in the action schemas. The identify of the action in error will
provide context for the subsequent refashioning of the referring
expression.\footnote {Another approach would be to use the
identity of the action in error to revise the beliefs that the agent
has attributed to the other conversant and to use the revised beliefs
in refashioning the plan. However, such reasoning is beyond the scope
of this work.}

%----------------------------------
\subsubsection*{Refashioning Plans}
%----------------------------------

If a conversant rejects a referring expression or postpones judgment
on it, then either the speaker or the hearer will refashion the
expression in the context of the rejection or postponement.  In
keeping with Clark and Wilkes-Gibbs, we use two discourse plans for
refashioning: \s{replace-plan} and \s{expand-plan}.  The first is
used to replace some of the actions in the referring expression plan
with new ones, and the second is to add new actions.  Replacements can
be used if the referring expression either overconstrains or
underconstrains the choice of referent, while the expansion can be
used only if it underconstrains the choice.  So, these plans can check
for these conditions.

The decomposition of the refashioning plans encode how a new referring
expression can be constructed from the old one.  This involves three
tasks: first, a single candidate referent is chosen; second, the
referring expression is refashioned; and third, this is communicated
to the hearer by way of the action \s{s-actions}, which was already
discussed.\footnote {Another approach would have been to separate the
communicative task from the first two 
\cite{lambert-carberry:91}.} The first step involves choosing a
candidate.  If the speaker of the refashioning is the agent who
initiated the referring expression, then this choice is obviously
pre-determined.  Otherwise, the speaker must choose the
candidate.  Goodman \shortcite{goodman:85} has addressed this problem
for the case of when the referring expression overconstrains the
choice of referent.  He uses heuristics to relax the constraints of
the description and to pick one that {\it nearly\/} fits it.  This problem
is beyond the scope of this paper, and so we choose one of the
referents arbitrarily (but see Heeman \shortcite{heeman:91:thesis} for
how a simplified version of Goodman's algorithm that relaxes only a
single constraint can be incorporated into the planning
paradigm).

The second step is to refashion the referring expression so that it
identifies the candidate chosen in the first step.  This is done by
using plan repair techniques \cite
{hayes:75:ijcai,wilensky:81:cbt,wilkens:85:ci}.  Our technique is to
remove the subplan rooted at the action in error and replan with
another action schema inserted in its place.  This technique has been
encoded into our refashioning plans, and so can be used for both
constructing repairs and inferring how another agent has repaired a
plan.
  
Now we consider the effect of these refashioning plans.  As we
mentioned in section~2, once the refashioning plan is accepted, the
common ground of the participants is updated with the new referring
expression.  So, the effect of the refashioning plans is that the
hearer will believe that the speaker wants the new referring
expression plan to replace the current one.  Note that this effect
does not make any claims about whether the new expression will in fact
enable the successful identification of the referent.  For if it did,
and if the new referring expression were invalid, this would imply
that the refashioning plan was also invalid, which is contrary to
Clark and Wilkes-Gibbs's model of the acceptance process.  So, the
understanding of a refashioning does not depend on the understanding
of the new proposed referring expression, but only on its derivation.

%==========================
\subsection{Action Schemas}
%==========================
 
This section presents action schemas for clarifications.  Each
clarification action includes a surface speech action in its
decomposition.  However, all reasoning is done at the level of the
clarification actions, and so the surface actions do not include any
constraints or effects.  The notation used in the action
schemas was given in Table~\ref{table:t3_table} above.

%---------------------------
\subsubsection*{accept-plan}
%---------------------------
\label{sec:accept-plan}

The discourse action \s{accept-plan}, shown in
Figure~\ref{plan:accept-plan}, is used by the speaker to establish the
mutual belief that a plan will achieve its goal.  The constraints of
the schema specify that the plan being accepted achieves its goal and
the decomposition is the surface speech action \s{s-accept}.  The
effect of the schema is that the hearer will believe that the speaker
has the goal that it be mutually believed that the plan achieves its
goal.
\begin{figure}[hbt]
\pplan{accept-plan}{\s{accept-plan} schema}{
  \planhead{    accept-plan(Plan) }
  \plancon{     bel(Speaker,achieve(Plan,Goal))}
  \plandec{     s-accept(Plan) }
  \planeff{     bel(Hearer,goal(Speaker,bel(Hearer,bel(Speaker, \\ &
                \plancont achieve(Plan,Goal)))))}
}
\end{figure}

%---------------------------
\subsubsection*{reject-plan}
%---------------------------
\label{sec:reject-plan}

The discourse action \s{reject-plan}, shown in
Figure~\ref{plan:reject-plan}, is used by the speaker if the referring
expression plan overconstrains the choice of referent.  The speaker
uses this schema in order to tell the hearer that the plan is invalid
and which action instance the evaluation failed in.  The constraints
require that the error occurred in an action instance whose yield
includes at least one primitive action.  The decomposition consists of
\s{s-reject}, which takes as its parameter the surface speech
actions that are in the yield of the problematic action.
\begin{figure}[hbt]
\pplan{reject-plan}{\s{reject-plan} schema}{
  \planhead{    reject-plan(Plan) }
  \plancon{     bel(Speaker,error(Plan,ErrorNode)) \\ &
                yield(Plan,ErrorNode,Acts) \\ &
                not(Acts = []) }
  \plandec{     s-reject(Plan,Acts) }
  \planeff{     bel(Hearer,goal(Speaker,bel(Hearer,bel(System, \\ &
                \plancont error(Plan,ErrorNode))))) }
}
\end{figure}

%-----------------------------
\subsubsection*{postpone-plan}
%-----------------------------
\label{sec:postpone-plan}

The schema for \s{postpone-plan}, shown in
Figure~\ref {plan:postpone-plan}, is similar to 
\s{reject-plan}.  However, it requires that the error in the
evaluation occurred in an action that does not decompose into any
primitive actions, which for referring expressions will be the
instance of \s{modifiers} that terminates the addition of
modifiers.
\begin{figure}[hbt]
\pplan{postpone-plan}{\s{postpone-plan} schema}{
  \planhead{    postpone-plan(Plan) }
  \plancon{     bel(Speaker,error(Plan,ErrorNode)) \\ &
                yield(Plan,ErrorNode,Acts) \\ &
                Acts = []}
  \plandec{     s-postpone(Plan,Acts) }
  \planeff{     bel(Hearer,goal(Speaker,bel(Hearer,bel(Speaker, \\ &
                 \plancont error(Plan,ErrorNode))))) }
}
\end{figure}

%----------------------------
\subsubsection*{replace-plan}
%----------------------------
\label{sec:replace-plan}

The \s{replace-plan} schema is used by the speaker to replace some
of the primitive actions in a plan with new actions.  Because we need
knowledge of the type of action where the error occurred in order that
we can refashion the invalid plan, the constraints of this schema are
more specific than those of the judgment plans.  The schema that we
give in Figure~\ref {plan:replace-plan}, for instance, is used to
refashion a referring expression plan in which the error occurred in
an instance of a \s{modifier} action.\footnote{If the error occurred
in an instance of \s{headnoun}, a different \s{replace-plan}
schema would need to be used, one that for instance relaxed the
category that was used in describing the object
\cite{goodman:85,heeman:91:thesis}.}
\begin{figure}[hbt]
\pplan{replace-plan}{\s{replace-plan} schema}{
  \planhead{    replace-plan(Plan) }
  \plancon{     bel(Speaker,error(Plan,ErrorNode)) \\ &
                content(Plan,ErrorNode,ErrorContent) \\ &
                ErrorContent = modifier(Entity,Object1,Cand,Cand1) }
  \plandec{     pick-one(Object,Cand) \\ &
                Replacement = modifier(Entity,Object,Cand,Cand2) \\ &
                substitute(Plan,Node,Replacement,NewPlan) \\ &
                replan(NewPlan,Acts) \\ &
                s-actions(Plan,Acts) }
  \planeff{     bel(Hearer,goal(Speaker,bel(Hearer,bel(Speaker, \\ &
                \plancont replace(Plan,NewPlan)))))}
}
\end{figure}

The decomposition of the schema specifies how a new referring
expression plan can be built.\footnote{We refer to the steps in the
decomposition that are not action headers as {\it mental actions\/}.
They need to be proved, just like constraints.} The first step, \s{pick-one(Object,Cand)}, chooses one of the objects that matched the
part of the description that preceded the error; if the speaker is not
the initiator of the referring expression, then this is an arbitrary
choice.  The second step specifies the header of the action schema
that will be used to replace the subplan that contained the error.
The third step substitutes the replacement into the referring
expression plan, undoing all variable instantiations in the old
plan.  This results in the partial plan \s{NewPlan}.  The
fourth step calls the plan constructor to complete the partial plan.
Finally, the fifth step is the surface speech action \s{s-actions},
which is used to inform the hearer of the surface speech actions that
are being added to the referring expression plan.

%---------------------------
\subsubsection*{expand-plan}
%---------------------------
\label{sec:expand-plan}

The \s{expand-plan} schema, shown in Figure~\ref
{plan:expand-plan}, is similar to the \s{replace-plan} schema shown
in Figure~\ref{plan:replace-plan}. The
difference is that instead of replacing one of the instances of \s{modifier}, it replaces the terminal instance of \s{modifiers} by a
\s{modifiers} subplan that distinguishes one of the objects from the
others that match, thus effecting an expansion of the surface speech
actions.  Even if the speaker thought that the referring expression as
it stands were adequate (since the candidate set \s{Cand} contains
only one member), she will construct a non-null expansion since the
replacement is the recursive version of \s{modifiers}.
\begin{figure}[htb]
\pplan{expand-plan}{\s{expand-plan} schema}{
  \planhead{    expand-plan(Plan) }
  \plancon{     bel(Speaker,error(Plan,ErrorNode)) \\ &
                content(Plan,ErrorNode,ErrorContent) \\ &
                ErrorContent = modifiers-terminate(Entity,Object1,Cand) }
  \plandec{     pick-one(Object,Cand) \\ &
                Replacement = modifiers-recurse(Entity,Object,Cand) \\ &
                substitute(Plan,ErrorNode,Replacement,NewPlan) \\ &
                replan(NewPlan,Acts) \\ &
                s-actions(Plan,Acts) }
  \planeff{     bel(Hearer,goal(Speaker,mb(Speaker,Hearer, \\ &
                \plancont replace(Plan,NewPlan))))}
}
\end{figure}

%================================================
\subsection{Plan Construction and Plan Inference}
%================================================

The general plan construction and plan inference processes are
essentially the same as those for referring expressions.  However, the
plan inference process has been augmented so as to embody the criteria
for understanding that were outlined in Section~\ref
{text:Planning-and-Clarifying}.  The inference of judgment plans must
be sensitive to the fact that such a plan includes the constraint that
the speaker found the judged plan to be in error even though the
hearer might not believe it to be.  So, the inference process is
allowed to assume that the speaker believes any constraint that the
goal of the plan implies.

In the case of a refashioning, the hearer might not view the proposed
referring expression plan as being sufficient for identifying the
referent, but would nonetheless understand the refashioning.  So, the
inference process requires only that the proposed referring expression
be derived---so that it can serve to replace the current plan---but
not that it be acceptable.  So, when a \s{replan} action is part of a plan that is being evaluated, the success
of this action depends only on whether the plan that is its parameter
can be derived, but not whether the derived plan is
valid.\footnote {Another approach would be to have the plan inference
process reason about the intended effects of the plan that it is
inferring in order to decide whether it should evaluate embedded plans
and whether this evaluation should affect the evaluation of the parent
plan.}

%===============================
\section{Modeling Collaboration}
%===============================

In the last two sections, we discussed how initial referring
expressions, judgments, and refashionings can be generated and
understood in our plan-based model.  In this section, we show how plan
construction and plan inference fit into a complete model of how an
agent collaborates in making a referring action successful.  Previous
natural language systems that use plans to account for the surface
speech acts underlying an utterance (such as Cohen and Perrault,
1979;\nocite {cohen-perrault:79} Allen and Perrault, 1980;\nocite
{allen-perrault:80} Appelt, 1985;\nocite {appelt:85j} Litman and
Allen, 1987\nocite {litman-allen:87}) model only the recognition or
only the construction of an agent's plans, and so do not address this
issue.

In order to model an agent's participation in a dialogue, we
need to model how the mental state of the agent changes as a result of
the contributions that are made to the dialogue.  The change in mental
state can be modeled by the beliefs and goals that a participant
adopts.  When a speaker produces an utterance, as long as the hearer
finds it coherent, he can add a belief that the speaker has made the
utterance to accomplish some communicative goal.  The hearer might
then adopt some goal of his own in response to this, and make an
utterance that he believes will achieve this goal.  Participants
expect each other to act in this way.  These social norms allow
participants to add to their common ground by adopting the inferences
about an utterance as mutual beliefs.

To account for how conversants collaborate in dialogue, however, this
co-operation is not strong enough.  Not only must participants form
mutual beliefs about what was said, they must also form mutual beliefs
about the adequacy of the plan for the task they are collaborating
upon.  If the plan is not adequate, then they must work together to
refashion it.  This level of co-operation is due to what Clark and
Wilkes-Gibbs refer to as a {\it mutual responsibility\/}, or what Searle
\shortcite {searle:90} refers to as a {\it we-intention\/}.  This allows
the agents to interact so that neither assumes control of the
dialogue, thus allowing both to contribute to the best of their
ability without being controlled or impeded by the other.  This is
different from what Grosz and Sidner \shortcite{grosz-sidner:90} have
called master-servant dialogues, which occur in teacher-apprentice or
information-seeking dialogues, in which one of the participants is
controlling the conversation (cf.~Walker and Whittaker, 1990\nocite
{walker-whittaker:90}).  Note that the non-controlling agent may be
helpful by anticipating obstacles in the plan \cite
{allen-perrault:80}, but this is not the same as collaborating.

The mutual responsibility that the agents share not only concerns the
goal they are trying to achieve, but also the plan that they are
currently considering.  This plan serves to coordinate their activity
and so agents will have intentions to keep this plan in their common
ground.  The plan might not be valid (unlike the {\it shared plan\/} of
\namecite {grosz-sidner:90}), so the agents might not mutually believe
that each action contributes to the goal of the plan.  Because of
this, agents will have a belief regarding the validity of the plan,
and an intention that this belief be mutually believed.

The discourse plans that we described in the previous section can now
be seen as plans that can be used to further the collaborative
activity.  Judgment plans express beliefs about the success of the
current plan, and refashioning plans update it.  So, the mental state
of an agent sanctions the adoption both of goals to express judgment
and of goals to refashion.  It also sanctions the adoption of
beliefs about the current plan.\footnote{The collaborative activity
also sanctions discourse expectations that the other participant's
utterances will pertain to the collaborative activity. We do not
explicitly address this however.} If it is mutually believed that one
of the conversants believes there is an error with the current plan,
the other also adopts this belief.  Likewise, if one of the
conversants proposes a replacement, the other accepts it.  Since both
conversants expect the other to behave in this way, each judgment
and refashioning, so long as they are understood, results in the
judgment or refashioning being mutually believed.  Thus the
current plan, through all of its refashionings, remains in the common
ground of the participants.

Below, we discuss the rules for updating the mental state after a
contribution is made.  We then give rules that account for the
collaborative process.\footnote{For simplicity, we represent the rules
for entering into a collaborative activity, adopting beliefs, and
adopting goals with the same operator, $\Longleftarrow$.  For a more
formal account, three different operators should be used.}

%===============================================
\subsection{Rules for Updating the Mental State}
%===============================================

After a plan has been contributed to the conversation, by way of its
surface speech actions, the speaker and hearer update their beliefs to
reflect the contribution that has been made.  Both assume that the
hearer is observant, can derive a coherent plan (not necessarily
valid), and can infer the communicative goal, which is expressed by
the effect of the top-level action in the plan.  We capture this by
having the agent that we are modeling, the system, adopt the belief
that it is mutually believed that the speaker intends to achieve the
goal by means of the plan.\footnote {See
\namecite{perrault:90} for how these inferences can be drawn by using
default rules.}
\infers{
bmb(system,user,plan(Speaker,Plan,Goal)) \\}
\noindent
The system will also add a belief about whether she believes the plan will
achieve the goal, and if not, the action that she believes to be in
error.  So, one of the following propositions will be adopted.
\infers{
bel(system,achieve(Plan,Goal)) \\
bel(system,error(Plan,Node)) \\}

After the above beliefs have been added, there are a number of
inferences that the agents can make and, in fact, can believe will be
made by the other participant as well, and so these inferences can be
mutually believed.  The first rule is that if it is mutually believed
that the speaker intends to achieve \s{Goal} by means of \s{Plan},
then it will be mutually believed that the speaker has \s{Goal} as
one of her goals.\footnote {All variables mentioned in the
rules are existentially quantified.}

\crule{bmb(system,user,goal(Agt1,Goal))}{
	bmb(system,user,plan(Agt1,Plan,Goal)) \ \& \\
        Agt1 $\in$ \{system,user\}}

The next rule concerns the adoption by the hearer of the intended
goals of communicative acts.  The communicative goal that we are
concerned with is where the speaker wants the hearer to believe that
the speaker believes some proposition.  This only requires that the
hearer believe the speaker to be sincere.  We assume that both
conversants are sincere, and so when such a communicative goal arises,
both participants will assume that the hearer has adopted the goal.
This is captured by rule (2).

\crule{bmb(system,user,bel(Agt1,Prop))}{
	bmb(system,user,goal(Agt1,bel(Agt2,bel(Agt1,Prop)))) \ \& \\
	Agt1,Agt2 $\in$ \{system,user\} \ \& \\
	not(Agt1 = Agt2)}

The last rule involves an inference that is not shared.  When the user
makes a contribution to a conversation, the system assumes that the
user believes that the plan will achieve its intended goal.

\crule{bel(system,bel(user,achieve(Plan,Goal)))}{
	bmb(system,user,plan(user,Plan,Goal))}

%======================================================
\subsection{Rules for Updating the Collaborative State}
%======================================================

The second set of rules that we give concern how the agents
update the collaborative state.  These rules have been revised from an
earlier version \cite{heeman:91:thesis} so as to better model the
acceptance process.

%------------------------------------------------------
\subsubsection{Entering into a Collaborative Activity}
%------------------------------------------------------

We need a rule that permits an agent to enter into a collaborative
activity.  We use the predicate \s{cstate} to represent that an agent
is in such a state, and this predicate takes as its parameters the
agents involved, the goal they are trying to achieve, and their
current plan.  Our view of when such a collaborative activity can be
entered is very simple: the system believes it is mutually believed
that one of them has a goal to refer and has a plan for doing so, but
one of them believes this plan to be in error.  The last part of the
condition states that if the speaker's referring expression was
successful from the beginning, no collaboration is necessary.  It is
not required that both participants mutually believe there is an
error. Rather, if either detects an error, then that conversant can
presuppose that they are collaborating, and make a judgment.  Once the
other recognizes the judgment that the plan is in error, the criteria
for him entering will be fulfilled for him as well.

\crule{cstate(system,user,Plan,Goal)}{
	bmb(system,user,goal(Agt1,Goal)) \ \& \\
	bmb(system,user,plan(Agt1,Plan,Goal)) \ \& \\
	Goal = knowref(Agt2,Agt1,Entity,Object) \ \& \\
	bel(system,bel(Agt3,error(Plan,Node))) \ \& \\
	Agt1,Agt2,Agt3 $\in$ \{system,user\} \ \& \\
	not(Agt1 = Agt2)
}

%------------------------------------------
\subsubsection{Adoption of Mutual Beliefs}
%------------------------------------------

In order to model how the state of the collaborative activity
progresses, we need to account for the mutual beliefs that
the agents adopt as a result of the utterances that are
made.

The first rule is for judgment moves in which the speaker finds the
current plan in error.  Given that the move is understood, both
conversants, by way of the rules given in section~5.1, will believe
that it is mutually believed that the speaker believes the current
plan to be in error.  In this case, the hearer, in the spirit of
collaboration, must accept the judgment and so also adopt the belief
that the plan is in error, even if he initially found the plan
adequate.  Since both conversants expect the hearer to behave in this
way, the belief that there is an error can be mutually believed.  Rule
(5), below, captures this.  (The adoption of this belief will cause
the retraction of any beliefs that the plan is adequate.)

\crule{bmb(system,user,error(Plan,Node))}{
	cstate(system,user,Plan,Goal) \ \& \\
	bmb(system,user,bel(Agt1,error(Plan,Node))) \ \& \\
	Agt1 $\in$ \{system,user\}
}

The second rule is for refashioning moves.  After such a move, the
conversants will believe it mutually believed that the speaker has a
replacement, \s{NewPlan}, for the current plan, \s{Plan}.  Again,
in the spirit of collaboration, the hearer must accept this
replacement, and since both expect each other to behave this way, both
adopt the belief that it is mutually believed that the new referring
expression plan replaces the old one.

\crule{bmb(system,user,replace(Plan,NewPlan))}{
	Agt1 $\in$ \{system,user\} \ \& \\
        cstate(system,user,Plan,Goal) \ \& \\
        bmb(system,user,error(Plan,Node)) \ \& \\
	bmb(system,user,bel(Agt1,replace(Plan,NewPlan)))
}

\noindent
In adopting this belief, the system updates the \s{cstate} by
replacing the current plan with the new plan, and adding beliefs that
capture the utterance of \s{NewPlan} as outlined in section~5.1
above.

The third rule is for judgment moves in which the speaker finds the
current plan acceptable.  Given that the move has been understood,
each conversant will believe it is mutually believed that the speaker
believes that the current plan will achieve the goal (second condition
of the rule).  However, in order to accept this move, each participant
also needs to believe that the hearer also finds the plan acceptable
(third condition).  This belief would have been inferred if it were
the hearer who had proposed the current plan, or the last
refashioning.  In this case, the speaker (of the acceptance) would
have inferred by way of rule (3) that the hearer believes the plan to
be valid; as for the hearer, given that he contributed the current
plan, he undoubtedly also believes it to be acceptable.

\crule{bmb(system,user,achieve(Plan,Goal))}{
        cstate(system,user,Plan,Goal) \ \& \\
        bmb(system,user,bel(Agt1,achieve(Plan,Goal)))  \ \& \\
	bel(system,bel(Agt2,achieve(Plan,Goal))) \ \& \\
	Agt1,Agt2 $\in$ \{system,user\} \ \& \\
        not(Agt1 = Agt2)
}

%------------------------------
\subsubsection{Adopting Goals}
%------------------------------

The last set of rules complete the circle.  They account for how
agents adopt goals to further the collaborative activity.  These goals
lead to judgment and refashioning moves, and so correspond to the
rules that we just gave for adopting mutual beliefs.

The first goal adoption rule is for informing the hearer that there is
an error in the current plan.  The conditions specify that \s{Plan} is
the current plan of a collaborative activity and that the speaker
believes that there is an error in it.

\crule{goal(system,bel(user,bel(system,error(Plan,Node))))}{
	cstate(system,user,Plan,Goal) \ \& \\
	bel(system,error(Plan,Node))
}

The second rule is used to adopt the goal of replacing the current
plan, \s{Plan}, if it has an error.  The rule requires that the
agent believe that it is mutually believed that there is an error in
the current plan.  So, this goal cannot be adopted before the goal of
expressing judgment has been planned.  Note that the consequent has an
unbound variable, \s{NewPlan}.  This variable will become bound when
the system develops a plan to achieve this goal, by using the action
schema \s{replace-plan} (Figure~\ref{plan:replace-plan} above).

\crule{goal(system,bel(user,bel(system,replace(Plan,NewPlan))))}{
        cstate(system,user,Plan,Goal) \ \& \\
        bmb(system,user,error(Plan,Node))
}

The third rule is used to adopt the goal of communicating the system's
acceptance of the current plan.  Not only must the system believe that
the plan achieves the goal, but it must also believe that the user
also believes this.  As mentioned above for rule (7), this last
condition prevents the system from trying to accept a plan that it has
itself just proposed. Rather, it can only try to accept a plan that
the other agent contributed, for it is just such plans for which it
will have the belief, by way of rule (3), that the user believes the
plan achieves the goal.

\crule{goal(system,bel(user,bel(system,achieve(Plan,Goal))))}{
	cstate(system,user,Plan,Goal) \ \& \\
	bel(system,achieve(Plan,Goal)) \ \& \\
	bel(system,bel(user,achieve(Plan,Goal)))
}

%==============================
\subsection{Applying the Rules}
%==============================

The rules that we have given are used to update the mental state of the
agent and to guide its activity.  Acting as the hearer, the system
performs plan inference on each set of actions that it observes, and
then applies any rules that it can.  When all of the
observed actions are processed, the system switches from the role of
hearer to speaker.

As the speaker, the system checks whether there is a goal that it can
try to achieve, and if so, constructs a plan to achieve it.  Next,
presupposing its partner's acceptance of the plan, it applies any
rules that it can.  It repeats this until there are no more goals.
The actions of the constructed plans form the response of the system;
in a complete natural language system, they would be converted to a
surface utterance.  The system then switches to the role of hearer.

%===================
\section{An Example}
%===================

We are now ready to illustrate our system in action.\footnote 
{The system is implemented in C-Prolog under Unix.}  For this
example, we use a simplified version
of a subdialogue from the London-Lund corpus
\cite[S.2.4a:1--8] {svartvik-quirk:80}:

\dialogue{
  \dialine{(6.1)}{A}{1}{See the weird creature.}
  \dialine{}{B}{2}{In the corner?}
  \dialine{}{A}{3}{No, on the television.}
  \dialine{}{B}{4}{Okay.}
}

\noindent The system will take the role of person B and we will give
it the belief that there are two objects that are ``weird''---a
television antenna, which is on the television, and a fern plant,
which is in the corner.

%------------------------------------------------
\subsection{Understanding ``The weird creature''}
%------------------------------------------------

\begin{figure}[t]
 \begin{center}
   \input{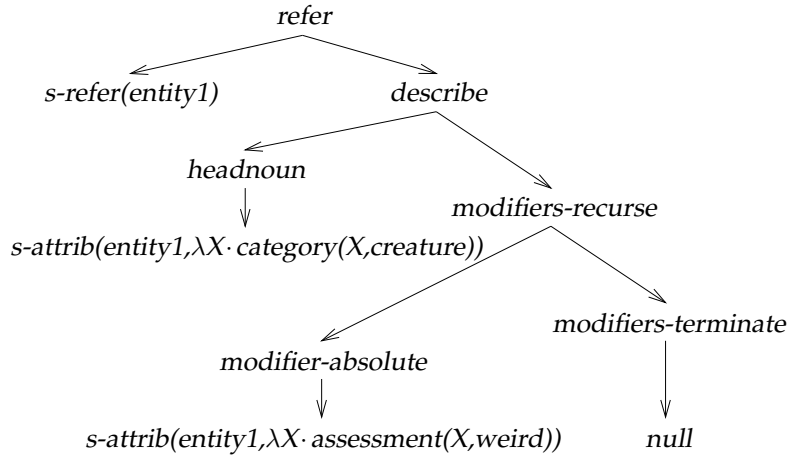}
 \end{center}
 \caption{\label{fig:t6_tree_1}Plan derivation (\s{p1}) for ``The weird creature''}
\end{figure}

For the first sentence, the system is given as input the surface
speech actions underlying ``the weird creature,'' as shown
below:
\infers{
s-refer(entity1) \\
s-attrib(entity1,\x{assessment(X,weird)}) \\
s-attrib(entity1,\x{category(X,creature)}) \\}
The system invokes the plan inference process, which finds
the plan derivations whose yield is the above set of surface speech
actions.  In this case, there is only one, and the system
labels it \s{p1}.  Figure~\ref{fig:t6_tree_1} shows the
derivation; arrows represent decomposition, and for brevity,
constraints and mental actions have been omitted and the parameters
only of the surface speech actions are shown.

Next, the plan derivation is evaluated.  The \s{subset}
constraint in the \s{headnoun} action is evaluated, which narrows
the candidate set to the antenna and the fern plant.  The \s{subset}
constraint in the \s{modifier} action is then evaluated,
which does not eliminate either of the candidates, since the system
finds both of them ``weird.''  The constraint on the \s{modifiers}
action that terminates the addition of modifiers is then
evaluated.  However, this constraint fails, since there are two
objects that match the description rather than one, as required.

The system then updates its beliefs.  As described in
section~5.1, the system adds the following beliefs to capture the
results of the plan inference process: that it is mutually believed
that the user has the goal of \s{knowref} and has adopted \s{p1}
as a means to achieve it, and that \s{p1} has an error on the
terminating instance of \s{modifiers}, node \s{p22}.
\infers{
\infer{bmb(system,user,plan(user,p1,knowref(system,user,entity1,Object)))}{a1} 
\infer{bel(system,error(p1,p22))}{a2}
}
The system next tries to apply the belief and goal adoption rules.
From rule (1) and belief (\ref{a1}), the system adds the belief that
it is mutually believed that the user has the goal that the system
\s{knowref} and from rule (3) and belief (\ref{a1}), it
adds the belief that the user believes that the plan achieves its goal.
\infers{
\infer{bmb(system,user,goal(user,knowref(system,user,entity1,Object)))}{a4}
\infer{bel(system,bel(user,achieve(p1,knowref(system,user,entity1,Object))))}{a3}}
Belief (\ref{a4}), along with (\ref{a1}) and (\ref{a2}), 
allows the system to apply rule (4), and so the system enters into a
collaborative activity, in which the goal is for it to know the
referent and in which the current plan is \s{p1}.
\infers{
\infer{cstate(system,user,p1,knowref(system,user,entity1,Object))}{a5}
}
Since the system believes there is an error in the current plan, it
applies rule (8), and so gives itself the communicative goal of
informing the user of the error in the current plan.
\infers{
\infer{goal(system,bel(user,bel(system,error(p1,p22))))}{a6}}

%-------------------------------------------
\subsection{Constructing ``In the corner?''}
%-------------------------------------------

Since there are no further belief or goal adoption rules that can be
applied, the system next checks for any goals that it can try to
achieve.  The only goal is (\ref{a6}), which is to inform the user of
the error in the plan.  Since the error in the referring plan is in
the terminating instance of \s{modifiers}, the plan constructor
builds an instance of \s{postpone-plan}, which it names \s{p26}.
(The schema was given in Figure~\ref {plan:postpone-plan}.)  
% New text:
Rather than realizing the surface speech action immediately, the
system plans ahead.  (This would allow an opportunistic process to
combine surface speech actions into a single utterance \cite
{appelt:85j}.)  So, the system, presupposing that the user understands
the system's plan, adds the following belief.
% Old text:
% To capture the utterance of the surface speech actions of the plan,
% the system, presupposing that the user understands them, adds the
% following belief.
%
\infers{
\infer{bmb(user,system,plan(system,p26,bel(user,bel(system,error(p1,p22)))))}{b1}}
It also adds the belief that this plan will achieve its goal.
\infers{
\infer{bel(system,achieve(p26,bel(user,bel(system,error(p1,p22)))))}{b2}}
Then by rule (1), the system adds the belief that it is mutually
believed that it has the goal.
\infers{
\infer{bmb(system,user,goal(system,bel(user,bel(system,error(p1,p22)))))}{b3}}
Then by rule (2), which captures the co-operativity of the agents in
communicative goals, it adds the belief that it is mutually believed
that the system believes there is an error.
\infers{
\infer{bmb(system,user,bel(system,error(p1,p22)))}{b4}}
Then, on the basis of (\ref{a5}) and (\ref{b4}) the system applies
rule (5), thus adopting the belief that it is mutually believed that
there is an error in the plan.  This presupposes the user's acceptance
of the judgment plan.
\infers{
\infer{bmb(system,user,error(p1,p22))}{b5}}
The system is now able to apply rule (9), on the basis of (\ref{a5})
and (\ref{b5}), and so adopts the goal of refashioning the invalid
referring expression plan and of informing the user of the new plan.
\infers{
\infer{goal(system,bel(user,bel(system,replace(p1,RPlan))))}{b6}}

\begin{figure}[t]
  \begin{center}\te
    \input{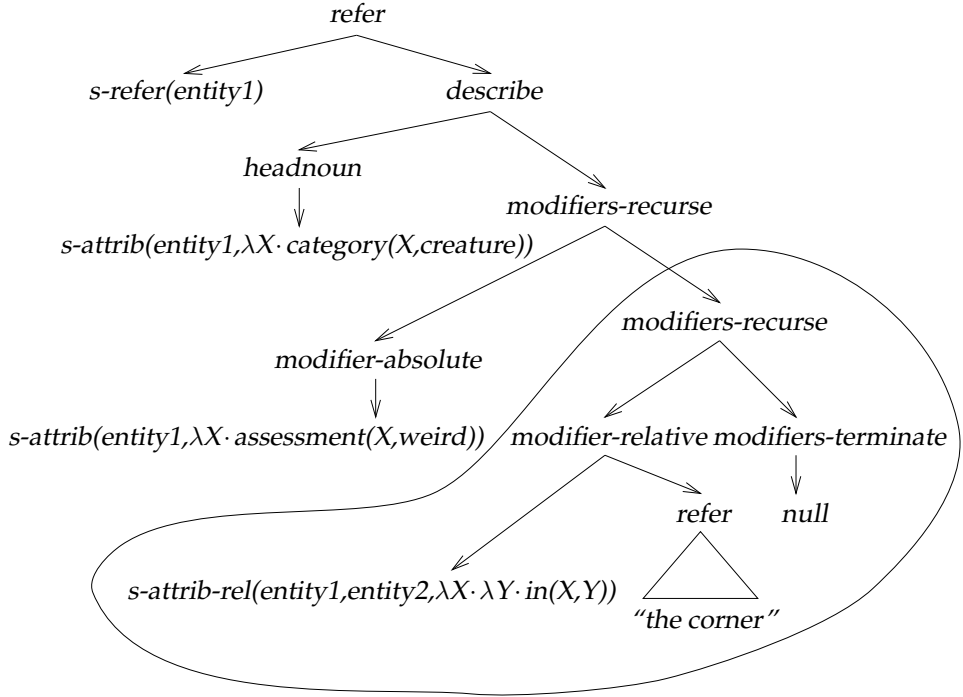}
  \end{center}
  \caption{\label{fig:t6_tree_2}Plan derivation (\s{p34}) for 
		``The weird creature in the corner''}
\end{figure}

Since no further rules can be applied, the system checks for goals
that it can try to fulfill, which will result in choosing (\ref{b6}).
To achieve this goal, the plan constructor builds an instance of
\s{expand-plan} (previously shown in Figure~\ref{plan:expand-plan}).
In doing this, the system chooses one of the objects that matched the
original description as the likely referent; in this case it happens
to choose the object in the corner, the fern plant, which the system
represents as \s{fern1}.  It then substitutes the \s{modifiers}
subplan that terminates the addition of modifiers with the header of
the \s{modifiers-recurse} action (with the chosen object
instantiated in).  The plan constructor is then called to fill in the
details, thereby creating the expansion. The expansion it chooses
includes a relative modifier (see Figure~\ref{plan:modifier-rel}) that
describes the object as being in the corner.  The new referring plan
(labeled \s{p34}) is shown in Figure~\ref{fig:t6_tree_2}, with the
expansion circled (we have abbreviated the derivation of ``the
corner'').  The surface speech action of \s{expand-plan} is
\s{s-actions}, which takes the surface speech actions of the
expansion, listed below, as its parameter.
\infers{
s-attrib-rel(entity1,entity2,\xy{in(X,Y)}) \\
s-refer(entity2) \\
s-attrib(entity2,\x{category(X,corner)}) \\
}

Next, the system assumes the user will understand the
refashioning, and, by way of rule (1) and (2), will be cooperative and
adopt the communicative goal that the system believes that the new
expanded plan replaces the old referring expression plan.  The end
result is given below as (\ref{c1}).
\infers{
\infer{bmb(system,user,bel(system,replace(p1,p34)))}{c1}}
The system, on the basis of (\ref{a5}) and (\ref{c1}), applies rule
(6), and so assumes that the user will accept the refashioning.  So,
the system adds the belief that it is mutually believed that the new
expanded plan replaces the old referring expression.
\infers{
\infer{bmb(system,user,replace(p1,p34))}{c2}}
This causes the belief module to update the current plan of the
collaborative activity (\ref{c7}).  Also, it adds the beliefs that
capture the utterance of the refashioned plan: that the system intends
it as a means to achieve the referring action and that it does achieve
this goal.\footnote{Even though the system has the referent
incorrectly identified in the goal of \s{knowref}, the goal itself is
still valid: for it to identify the referent corresponding to
\s{entity1}.}
\infers{
\infer{cstate(system,user,p34,knowref(system,user,entity1,Object)))}{c7}
\infer{bmb(system,user,plan(system,p34,knowref(system,user,entity1,fern1)))}{c5}
\infer{bel(system,achieve(p34,knowref(system,user,entity1,fern1)))}{c6}}

The two plans that were constructed, \s{postpone-plan} and
\s{expand-plan}, give rise to the output of the surface speech
actions \s{s-postpone} and \s{s-expand}, which would be realized as
``in the corner?''.\footnote{Although our model does not account for
the questioning intonation, it could be a manifestation of the
\s{s-postpone}.}

%===================================================
\subsection{Understanding ``No, on the television''}
%===================================================

The user next utters ``No, on the television.''  This would get parsed
into two separate surface speech actions, an \s{s-reject}
corresponding to ``no'', and an \s{s-actions} corresponding to ``on
the television.''  For simplicity, the plan inference process is
invoked separately on each.

The system starts with the \s{s-reject} action.  We assume that the
parser can determine from context that the ``no'' is rejecting the
surface speech actions that were previously added and so the parameter
of \s{s-reject} is a list of these actions.  From this, it derives a
plan whose yield is the \s{s-reject} action, and this plan is an
instance of \s{reject-plan} (previously shown in
Figure~\ref{plan:reject-plan}).  The system then evaluates the constraints of the
plan, which results in it determining which action in the plan the
user found to be in error.  This is done by evaluating the constraints
of \s{reject-plan}, and so finding the action whose yield is the
surface speech actions that were rejected.  This will be \s{p56},
the \s{modifiers-relative} action that described the object as being
in the corner.  The resulting belief, after applying rules (1) and
(2), is the following.
\infers{
\infer{bmb(system,user,bel(user,error(p34,p56)))}{d1}}
The system then applies the appropriate acceptance rule, rule (5), and
so adopts the belief that the error is mutually believed.
\infers{
\infer{bmb(system,user,error(p34,p56))}{d2}}
With this belief, the system will have the context that it needs to
understand the user's refashioning plan.

The system next performs plan recognition starting with the second
surface speech action, \s{s-actions}, which corresponds to the
refashioning ``on the television''.  It takes as a parameter the
following list of actions:\footnote{We assume that the parser
determines the appropriate discourse entities in these actions:
\s{entity1} is the discourse entity for the object being referred to,
and
\s{entity3} is another discourse entity.}
\infers{
s-attrib-rel(entity1,entity3,\xy{on(X,Y)}) \\
s-refer(entity3) \\
s-attrib(entity3,\x{category(X,television)}) \\}
The system finds two plan derivations that account for the primitive
action, one an instance of \s{replace-plan} (see Figure~\ref
{plan:replace-plan}) and the other an instance of \s{expand-plan}
(Figure~\ref{plan:expand-plan}).  Next it evaluates the constraints of
each derivation.  The constraints of \s{expand-plan} do not hold since
the action in error, \s{p56}, is not an instance of
\s{modifiers-terminate}, so this plan is eliminated.  The constraints
(and mental actions) of \s{replace-plan} do hold, and so the system
is able to derive the refashioned referring plan, which it labels
\s{p104}.

Since this instance of \s{replace-plan}
is the only valid derivation corresponding to the surface speech
actions observed, the system takes it as the plan behind the user's
utterance.  As a result, the system adds the following belief (after
applying rule (1) and (2)).
\infers{
\infer {bmb(system,user,bel(user,replace(p34,p104)))}{e1}}
The system then applies the acceptance rule for refashioning plans,
rule (6), and so adopts the refashioning as mutually believed.
\infers{
\infer {bmb(system,user,replace(p34,p104))}{e2}}
This causes the belief module to update the current plan of the
collaborative activity and to add the belief that the user contributed
the new referring expression plan.
\infers{
\infer{cstate(system,user,p104,knowref(system,user,entity1,Object))}{e3}
\infer
{bmb(system,user,plan(user,p104,knowref(system,user,entity1,antenna1))}{e3.5}}
The new referring plan will already have been evaluated. The subplan
corresponding to ``the television'' would have been understood without
problem,\footnote {If ``the television'' is not understood, then since
it is a referring expression in its own right, the conversants could
collaborate on identifying its referent independently of the referent
of ``the weird creature''; that is, the participants could enter into
an embedded collaborative activity by focusing on one part of the
current plan.} and the modifier corresponding to ``on the television''
would have narrowed down the candidates that matched ``weird
creature'' to a single object, \s{antenna1}.  So, the belief module
adds the belief that the system finds the new referring plan to be
valid.  Also, by way of rule (3), the system adds the belief that
the user also does, since the user had proposed it.
\infers{
\infer{bel(system,achieve(p104,knowref(system,user,entity1,antenna1)))}{e4}
\infer{bel(system,bel(user,achieve(p104,knowref(system,user,entity1,antenna1))))}{e5}}

%---------------------------------
\subsection{Constructing ``Okay''}
%---------------------------------

On the basis of (\ref{e3}), (\ref{e4}), and (\ref{e5}), the system is
able to apply rule (10), and so adopts the goal of accepting the
plan.
\infers{
\infer{goal(system,bel(user,bel(system,achieve(p104, \\
  \hspace*{1em} knowref(system,user,entity1,antenna1))))))}{f1}}
The plan constructor achieves this by planning an instance of
\s{accept-plan}, which results in the surface speech action 
\s{s-accept}, which would be realized as ``Okay.''  Then, after the
application of rules (1), (2), and most importantly (7), the system
adopts the belief that it is mutually believed that the plan achieves
the goal of referring.
\infers{
\infer{bmb(system,user,achieve(p104,knowref(system,user,entity1,antenna1)))}{f2}}

%====================================
\section{Comparisons to Related Work}
%====================================

In providing a computational model of how agents collaborate upon
referring expressions, we have touched on several different areas of
research.  First, our work has built on previous work in referring
expressions, especially their incorporation into a model based on the
planning paradigm.  Second, our work has built on the research done in
modeling clarifications in the planning paradigm and on plan repair.
Third, our work is related to the research being done on modeling
collaborative and joint activity.

%---------------------------------
\subsection{Referring Expressions}
%---------------------------------

Cohen \shortcite{cohen:81} and Appelt \shortcite{appelt:85j} have also
addressed the generation of referring expressions in the planning
paradigm.  They have integrated this into a model of generating
utterances, a step that we haven't taken.  However, we have extended
their model by incorporating even the generation of the components of
the description into our planning model.  One result of this is that
our surface speech actions are much more fine-grained.

%==========================================
\subsection{Clarifications and Plan Repair}
%==========================================

An important part of our work involves accounting for clarifications
of referring expressions by using meta-actions that incorporate plan
repair techniques.  This approach is based on Litman and Allen's work
\shortcite {litman-allen:87} on understanding clarification
subdialogues, in which meta-actions were used to model discourse
relations, such as clarifications.  There are several major
differences between our work and theirs.  First, our work addresses
not only understanding but also generation and how these two tasks fit
into a model of how agents collaborate in discourse.  Second, Litman
and Allen use a stack of unchanging plans to represent the state of
the discourse.  We, however, use a single {\it current plan\/}, modifying
it as clarifications are made.  This difference has an important
ramification, for it results in different interpretations of the
discourse structure.  Consider dialogue (7.1), which was collected at
an information booth in a Toronto train station \cite{horrigan:77}.
(Although the participants are not collaborating in making a referring
expression, the dialogue will serve to illustrate our point.)

\dialogue{
  \dialine{(7.1)}{P}{1}{The 8:50 to Montreal?}
  \dialine{}{C}{2}{8:50 to Montreal.  Gate 7.}
  \dialine{}{P}{3}{Where is it?}
  \dialine{}{C}{4}{Down this way to your left.  Second one on the left.}
  \dialine{}{P}{5}{OK.  Thank you.}
}

\noindent Litman and Allen represent the state of the discourse after
the second utterance as a clarification
of the passenger's \s{take-train-trip} plan.  The information
that the train boards at gate 7 is represented only in the
clarification plan.  So, when the passenger asks ``Where is it?'',
their system, acting as the clerk, cannot interpret this as a
clarification of the \s{take-train-trip} plan, since the utterance
``cannot be seen as a step of [that] plan'' (p.~188).  So, it is
interpreted instead as a request for a clarification of the clerk's
``Gate 7'' response, implicitly assuming that ``Gate 7'' was not
accepted.  In our model, the acceptance of ``Gate 7'' would be
presupposed, and so it would be incorporated into the
\s{take-train-trip} plan.  So, the passenger's question of ``Where is
it?'' would be viewed as a request for the clerk to clarify that plan.

The work of Moore and Swartout \shortcite {moore-swartout:91:nlgaicl},
Cawsey \shortcite {cawsey:91:aaai}, and Carletta \shortcite
{carletta:91:tr-a} on interactive explanations also addresses
clarifications using plan repair techniques.  This body of work uses
plan construction techniques to generate explanations, and uses the
constructed plan as a basis for recovery strategies if the user
doesn't understand the explanation.  In the cases of Cawsey and
Carletta, both use meta-actions to encode the plan repair techniques.
However, none of these approaches are within a collaborative
framework, in which either agent can contribute to the development of
the plan.

Other relevant work is that of Lambert and Carberry \shortcite
{lambert-carberry:91}.  In their model of understanding
information-seeking dialogues, they propose a distinction between
problem-solving activities and discourse activities.  In contrast, our
clarifications embody both functions in the same actions, thus
allowing for a simpler approach to inferring the refashioned referring
expressions, since we need not chain to a meta-operator.  In later
work, \namecite{chucarrol-carberry:94:aaai} extended this model to
generate responses to proposals that are viewed as sub-optimal or
invalid.  Like Litman and Allen \shortcite{litman-allen:87}, they
adopt the view that subsequent modifications apply to the preceding
modification, rather than the underlying plan.

%=========================
\subsection{Collaboration}
%=========================

Grosz, Sidner, and Lochbaum \cite
{grosz-sidner:90,lochbaum-grosz-sidner:90} are interested in the type
of plans that underlie discourse in which the agents are collaborating
in order to achieve some goal.  They propose that agents are building
a {\it shared plan\/} in which participants have a collection of beliefs
and intentions about the actions in the plan.  Our model differs from
theirs in two important aspects.  First, not only do agents have a
collection of beliefs and intentions regarding the actions of a shared
plan, we feel that they also have an intention about the goal \cite
{searle:90,cohen-levesque:91:ijcai}.  It is this intention, in
conjunction with the current plan, that sanctions the adoption of
beliefs and intentions about potential actions that will contribute to
the goal, rather than just the shared plan.

Second, we feel that their definition of a partial shared plan is too
restrictive.  Although they address partial plans, they require, in
order for an action to be part of a partial shared plan, that both
agents believe that the action {\it contributes \/} to the goal.  However,
this is too strong.  In collaborating to achieve a mutual goal,
participants sometimes propose an action that is not believed by the
other participant or even by the participant that is proposing it.  In
failing to represent such states, their model is unable to
represent the intermediate states in which a hearer might have
understood how the speaker's utterance contributes to a plan, but
doesn't agree with it.  This is important, since if the refashioned
plan is invalid, only the referring expression should be
refashioned, not the refashioning itself.

Traum (1991;\nocite{traum:91:proposal} Traum and Hinkelman, 1992\nocite{traum-hinkelman:92:ci}) is concerned with providing a
computational model of {\it grounding\/}, the process in which
conversational participants add to the common ground of a conversation
\cite {clark-schaefer:89,clark-brennan:90}.  Traum models the
grounding process by proposing that utterances move through a number
of states, `pushed' by grounding acts, which include initiate,
continue, repair, request repair, acknowledge, and request
acknowledge.  Once an utterance has been acknowledged, it will reside
in mutual belief as a proposal of the person who initiated it.  The
proposal state is a subspace of the mutual belief space of the
conversants.  Only once it has been accepted, will it be moved into
the {\it shared\/} space (also in mutual belief).  Unlike Traum's, our
work does not differentiate the proposal state from the shared state.
If a proposal is understood, it is incorporated into the current plan.
Judgments of acceptability are not on proposals but on the current
plan, or a part of it.

Sidner \shortcite {sidner:92:aaai} addressed the issue of how
conversational participants collaborate in building a shared plan.  In
this work, Sidner presents a number of speech actions for use in
collaborative tasks.  These actions are those that an artificial agent
could use in negotiating which actions or beliefs to accept into the
shared plan of the agents.  As with Traum, it is the {\it proposals\/}
that are refashioned, before they are integrated into the shared plan,
rather than the shared plan.

Cohen and Levesque \shortcite{cohen-levesque:91:ijcai} focus on
formalizing joint intention in a logic.  They use this formalism to
explain how such elements of communication as confirmations arise when
agents are engaging in a joint action.  However, they have not
addressed how agents collaborate in building a plan, only how agents
collaborate while executing a plan.  Once this limitation is overcome,
their approach could offer us a route for formalizing the mental
states of the collaborating agents in our model and for proving that
our acceptance and goal adoption rules
follow from such states.

%===================
\section{Conclusion}
%===================

We have presented a computational model of how a conversational
participant collaborates in making and understanding a referring
expression, based on the view that language is goal-oriented
behavior.  This has allowed us to do the following.  First, we have accounted
for the tasks of building a referring expression and identifying its
referent by using plan construction and plan inference.  Second, we
have accounted for the conversational moves that participants make
during the acceptance process by using meta-actions.  Third, we have
accounted for collaborative activity by proposing that agents are in a
certain mental state that includes a goal, a plan that they are
currently considering, and intentions.  This mental state sanctions
the acceptance of clarification plans, and sanctions the adoption of
goals to clarify.  Although our work has focused on referring
expressions, we feel that it is relevant to collaboration in general
and to how agents contribute to discourse.

This paper is based on the model of collaboration proposed by Clark
and Wilkes-Gibbs \shortcite {clark-wilkesgibbs:86}.  Their model makes
two strong claims about how agents collaborate.  First, it minimizes
the distinction between the roles of the person who initiates the
referring expression and the person who is trying to identify it.
Both have the same moves available to them, for either can judge the
description and either can refashion it.  This allows both
participants to contribute without being controlled or impeded by the
other.  Second, their model gives special status to the role of the
current referring expression (current plan): participants judge and
refashion the current referring expression directly, rather than
recursively modifying modifications
(e.g.~Litman and Allen, 1987; Chu-Carrol and Carberry, 1994\nocite{litman-allen:87,chucarrol-carberry:94:aaai}) or
incrementally adding to the current plan with each accepted proposal
(e.g.~Traum and Hinkelman, 1992; Sidner, 1992\nocite{traum-hinkelman:92:ci,sidner:92:aaai}).  In our work, we
have taken Clark and Wilkes-Gibbs's descriptive model and recast it
into a computational one, thus demonstrating the computational
feasibility of their work and its compatibility with current practices
in artificial intelligence.

There are many ways that this research could be extended.  Perhaps the
most obvious would be to extend the planning component of our model.
First, our coverage of referring expressions could be extended to
handle references to objects in focus and to descriptions that include
a plan of physical actions for identifying the referent.  Second, the
treatment of clarifications could be improved; specifically, how plan
failures are reasoned about, how plan failures affect the agent's
beliefs, and how these failures are repaired.  Third, this research
needs to be integrated into a more complete plan-based approach to
language, and needs to be extended so as to handle more general
discourse plan failures
\cite{mcroy-hirst:93:eacl,mcroy-hirst:94,horton-hirst:91:dsnlug,heeman:93:aaaiss,edmonds:94:coling,hirst-etal:94:sc}.
A benchmark for such future work could be dialogue (8.1) below, from
the London-Lund corpus \cite [S.2.4a:1--8] {svartvik-quirk:80}, which
is the basis of the example used in section 6.  This dialogue shows
how collaboration on a referring expression can be embedded in other
activities, how agents can return back to a collaborative activity,
and even how agents can take advantage of a mistaken referent.

\dialogue{
  \dialine{(8.1)}{A}{1}{What's that weird creature over there?}
  \dialine{}{B}{2}{In the corner?}
  \dialine{}{A}{3}{\it affirmative noise}
  \dialine{}{B}{4}{It's just a fern plant.}
  \dialine{}{A}{5}{No, the one to the left of it.}
  \dialine{}{B}{6}{That's the television aerial.  It pulls out.}
}

A second avenue for future work is to further investigate
collaborative behavior and protocols for interaction.  We need
to formalize what it means for agents to be collaborating, in a theory
that takes account of rational interaction and the beliefs and
knowledge of the participants.  Such a theory would do the following.
First, it would give a more complete motivation for the processing
rules that we used for how agents interact in a collaborative
activity.  Second, it would account for why agents would enter into
such a mode of interaction, how it is initiated, how it is carried
forward (especially how agents' beliefs and knowledge influence their
actions), and how it ends.  Third, it would be extendable to other
forms of interaction, such as information-seeking dialogues.  Fourth,
it would specify how collaborative activity could be embedded in, or
embed, other types of interactions.  By answering these questions, we
will not only have a better model to base natural language interfaces
on, but we will also have a better understanding of how people
interact.

%-------------------------
\section*{Acknowledgments}
%-------------------------

We would like to thank James Allen, Hector Levesque, and the referees
at {\it Computational Linguistics} for their comments on an earlier
version of this paper.  We would also like to especially thank Janyce
Wiebe for her invaluable contribution to the development of this work.
As well, we are grateful for comments from, and discussions with,
Diane Horton, Susan McRoy, Massimo Poesio, and David Traum.  Funding
at the University of Toronto and the University of Rochester was
provided by the Natural Sciences and Engineering Research Council of
Canada, with additional funding at Rochester provided by NSF under
Grant IRI-90-13160 and ONR/DARPA under Grant N00014-92-J-1512.

%===========================
\bibliographystyle{fullname}
%===========================
%\bibliography{/u/heeman/bib/abbrev,/u/heeman/bib/heeman,/u/heeman/bib/nlu,/u/heeman/bib/plan,/u/heeman/bib/misc,/u/heeman/bib/books,/u/heeman/bib/cross}

\end{document}